\documentclass[final,12pt]{alt2024} % Include author names

\title[Optimal Regret in Collaborative Bandits]{Optimal Regret Bounds for Collaborative Learning in Bandits}
\usepackage[utf8]{inputenc} % allow utf-8 input
\usepackage[T1]{fontenc}    % use 8-bit T1 fonts
\usepackage{hyperref}       % hyperlinks
\usepackage{url}            % simple URL typesetting
\usepackage{booktabs}       % professional-quality tables
\usepackage{amsfonts}       % blackboard math symbols
\usepackage{nicefrac}       % compact symbols for 1/2, etc.
\usepackage{microtype}      % microtypography
\usepackage{xcolor}         % colors
\usepackage{mathtools}
\usepackage{algorithm}
\usepackage{algorithmic}
\usepackage{times}
\altauthor{%
 \Name{Amitis Shidani} \Email{amitis.shidani@stats.ox.ac.uk}\\
 \addr University of Oxford
 \AND
 \Name{Sattar Vakili} \Email{sv388@cornell.edu}\\
 \addr MediaTek Research%
}

\usepackage{url}            % simple URL typesetting
\usepackage{booktabs}       % professional-quality tables
\usepackage{amsfonts}       % blackboard math symbols
\usepackage{nicefrac}       % compact symbols for 1/2, etc.
\usepackage{microtype}      % microtypography
\usepackage{mathtools}
\usepackage{times}

\def\nn{\nonumber}
\DeclareMathOperator*{\argmax}{arg\,max}
\DeclareMathOperator*{\argmin}{arg\,min}

\def\Nn{\mathbb{N}}
\def\E{\mathbb{E}}
\def\Pr{\mathbb{P}}
\def\Rr{\mathbb{R}}

\def\EB{\mathcal{B}}
\def\EE{\mathcal{E}}

\def\Rc{\mathcal{R}}
\def\Pc{\mathcal{P}}
\def\Oc{\mathcal{O}}
\def\Fc{\mathcal{F}}
\def\Tc{\mathcal{T}}
\def\Cc{\mathcal{C}}

\def\one{\mathbf{1}}

\def\Deltah{\hat{\Delta}}
\def\Deltat{\tilde{\Delta}}

\def\alg{\mbox{CExp$^2$}}
\def\eg{\emph{e.g.}}

\def\ie{\text{ie}}
\def\ge{\text{ge}}
\def\ex{\text{ex}}

\begin{document}

\maketitle

\begin{abstract}
We consider regret minimization in a general collaborative multi-agent multi-armed bandit model, in which each agent faces a finite set of arms and may communicate with other agents through a central controller. The optimal arm for each agent in this model is the arm with the largest expected \emph{mixed} reward, where the mixed reward of each arm is a weighted average of its rewards across all agents, making communication among agents crucial. While near-optimal sample complexities for best arm identification are known under this collaborative model, the question of optimal regret remains open. In this work, we address this problem and propose the first algorithm with order optimal regret bounds under this collaborative bandit model. Furthermore, we show that only a small constant number of expected communication rounds is needed.
\end{abstract}


\section{Introduction}
In the classic multi-armed bandit problem~\citep{LaiRobbins85bandits, auer2002finite} 
an agent sequentially interacts with a set of arms. At each time, the agent selects one of the arms and receives the corresponding reward drawn from an unknown distribution. The goal of the agent is to minimize \emph{regret} defined as the total loss in the rewards compared to always selecting the arm with the highest expected reward. The trade-off between \emph{exploration} ---selecting each arm several times to learn the unknown distributions--- and \emph{exploitation} ---capitalizing on the information gathered so far to select the arm with the highest reward--- is central to designing efficient bandit algorithms that minimize the regret. The bandit problem has also been studied under a different setting, referred to as the best arm identification, where the goal of the agent is to identify the optimal arm with high probability with as a small number of samples as possible~\citep{EvenDaral06, Jamiesonal14LILUCB}.

There is an increasing interest in collaborative learning due to modern applications in traffic routing, communication systems, social dynamics, federated learning, and the Internet of Things (IoT). We consider a statistical learning point of view to collaborative learning through a general multi-agent multi-armed bandit model. This collaborative bandit model was proposed in~\cite{reda2022collaborative}, which itself was a generalization of the federated learning with the personalization model introduced in~\cite{shi2021federated}.
%The collaborative bandit model studied in this work was described in~\cite{reda2022collaborative}.
In this model, $M$ agents interact with $K$ arms. An agent $m\in[M]$ observes a \emph{local} reward $X_{k,m}(t)$ for playing arm $k\in[K]$ at time $t\in[T]$. The actual reward that the agent receives, however, is a \emph{mixed} reward $X'_{k,m}(t)$ that is a weighted average of local rewards over all $M$ agents for the same arm $k$. A special case of this collaborative bandit model limited to particular weights was introduced in \cite{shi2021federated} for a federated learning setting. A detailed problem formulation highlighting the special case of \cite{shi2021federated} is given in Section~\ref{sec:PF}.

The interdependence among agents due to mixed rewards naturally makes \emph{communication} among agents crucial. In fact, it is easy to design problem instances where without communication a trivial linear in $T$ regret is inevitable. The agents in the collaborative bandit model are allowed to communicate through a central controller or server. Specifically, the agents can send information (\eg, empirical means of past local observations) to the server, which will broadcast this information to all other agents. An ideal algorithm obtains a good performance in terms of regret 
%(or sample complexity in best arm identification) 
with few communication rounds. 

The realistic scenarios of collaborative bandit include applications to recommendation systems, for which one may want to go beyond uniformly personalized learning; favoring the local rewards of agent $m$ over other agents’ observations in the identification of the optimal arm~\citep[][]{shi2021federated}. The setting is also suitable for adaptive clinical trials on $K$ therapies, run by $M$ teams who have access to different sub-populations of patients. In this context, each sub-population typically aims to find its 
%(local) 
best treatment~\citep[see details in][]{reda2022collaborative}. Similarly, this setting has applications in routing and communication systems~\citep[][]{shi2021federated}.

\subsection{Related Work and Contribution}

The authors in \cite{shi2021federated} provided a conjecture on the lower bound on regret under the special case of federated learning with personalization\footnote{The term personalization here emphasizes the property of this collaborative learning model that the mixed reward distributions are agent specific, in contrast to settings where all agents share the same global reward model~\citep[see, \eg, ][]{tao2019collaborative, shi2021global}.}.
They also introduced \emph{personalized federated upper confidence bound} (PF-UCB) algorithm that achieves an $\Oc(\log(T))$ regret where the implied constant in the $\Oc$ notation is suboptimal in terms of problem parameters.
The authors in \cite{reda2022collaborative} primarily studied best arm identification under the collaborative bandit model. They proved an $\Omega(s^*\log(\frac{1}{\delta}))$ information-theoretic lower bound on the sample complexity for the event that, with probability at least $1-\delta$, each agent identifies their own best arm. In this lower bound, $s^*$ is the solution to a constrained optimization problem involving the weights and the mixed reward gaps. In addition, they used the knowledge of the lower bound to propose \emph{weighted collaborative phased elimination for best arm identification} (W-CPE-BAI) algorithm, achieving near-optimal performance $\Oc(s^*\log(\frac{1}{\Delta'_{\min}})\log(\frac{1}{\delta}))$, where $\Delta'_{\min}$ is the smallest gap in the expected mixed rewards of the best arm and a suboptimal arm across all agents. 

In the context of our work, \cite{reda2022collaborative} established an $\Omega(c^*\log(T))$ lower bound on regret
%---in contrast to sample complexity in the best arm identification setting--- 
that also confirmed the conjecture in \cite{shi2021federated}. 
Similar to the previous case, in this lower bound, $c^*$ is the solution to a constrained optimization problem involving the weights and the mixed reward gaps (see Lemma~\ref{lem:lowerbound} for details). However, designing an algorithm achieving near-optimal regret was proven to be a challenging task and remained as an open problem.
%~\citep{reda2022collaborative}. 
The difficulty arises from the structure of the constrained optimization problem associated with $c^*$, which is essentially different from the one associated with~$s^*$. As noted in~\cite{reda2022collaborative}, a standard analysis of phased elimination algorithms would lead to an $\Oc(\frac{c^*}{\Delta'_{\min}}\log(T))$ regret bound. In this case, the $\frac{1}{\Delta'_{\min}}$ gap is exponentially worse than the $\log(\frac{1}{\Delta'_{\min}})$ gap in the previous case, and can result in poor performance guarantees.

In this work, we present a novel regret minimization algorithm called \emph{collaborative double exploration} (\alg) and demonstrate that it achieves an $\Oc(c^*\log(T))$ regret, with only absolute constants hidden in the $\Oc$ notation. Our regret guarantees for this algorithm match the problem-specific regret lower bound given in~\cite{reda2022collaborative}.
\alg~comprises a sub-logarithmic initial round-robin exploration phase, followed by a guided exploration phase that is designed based on the constrained optimization problem associated with $c^*$ (hence, the name double exploration). We prove that by utilizing the initial exploration phase, the guided exploration phase can gather enough information to efficiently exploit the best arm, resulting in an optimally bounded total regret.
%(up to absolute constants). 
Moreover, we show that \alg~achieves the order optimal regret bound with only a constant number of expected communication rounds. Specifically, the number of communication rounds is 
%bounded by a constant 
$\Oc(\log(\frac{1}{\Delta_{\min}}))$. Furthermore, with probability at least $1-\frac{1}{\log(T)}$, there are only $2$ communication rounds in \alg, implying that the expected number of communication rounds is only $2$ as~$T$ grows large. 

\subsection{Paper Structure}

In Section~\ref{sec:PF}, we introduce the notation and provide the details of the collaborative bandit model. We also overview the background on regret lower bound and confidence intervals in this setting. In Section~\ref{sec:alg}, we describe our collaborative bandit algorithm \alg. The analysis of the algorithms is given in Section~\ref{sec:anal}, while some details are deferred to the appendix. Further related work is discussed in Appendix~\ref{appx:rw}.

\section{Problem formulation}\label{sec:PF}

In the collaborative bandit model, there are $K$ arms and $M$ agents. At each sequential time $t\in [T]$, $T\in\Nn$, each agent $m\in[M]$ is allowed to select an arm $k\in[K]$ and observe the corresponding local reward $X_{k,m}(t)$ independent across $k,m$ and independent and identically distributed across $t$. We use the notation $[N]=\{1,2,\dots,N\}$, for $N\in\Nn$, throughout the paper. The random local reward $X_{k,m}(t)$ is distributed according to a $\sigma$-sub-Gaussian distribution~$\nu_{k,m}$. 
We say that the distribution of a random variable $X$ is $\sigma$-sub-Gaussian when for all $\eta\in\Rr$, $\E[\exp(\eta (X-\E[X]))]\le \exp(\frac{\eta^2\sigma^2}{2})$, for some $\sigma>0$. This property allows us to use statistical confidence bounds on the mean rewards in algorithm design.   
We use $\mu_{k,m}=\E[X_{k,m}(t)]$ to denote the expected value of the rewards. The agents are interested in the mixed rewards of the arms defined using a fixed and known weight matrix $W=(w_{n,m})_{n,m}\in [0,1]^{M\times M}$, satisfying $\sum_{n\in[M]}w_{n,m}=1$ for all $m\in[M]$, i.e.\ the columns of~$W$ add up to~$1$. The weight $w_{n,m}$ represents the importance weight of agent $n$ for agent $m$. Specifically, the mixed reward of agent $m$ from arm $k$ at time $t$ is defined as a weighted average of the local rewards over all agents
\begin{equation}
     X'_{k,m}(t)=\sum_{n=1}^Mw_{n,m}X_{k,n}(t),
\end{equation}
and its expectation, referred to as expected mixed reward, is denoted by
$
\textstyle \mu'_{k,m} = \sum_{n=1}^Mw_{n,m}\mu_{k,n}
$.
We use $k^*_m=\arg\max_{k\in[K]}\mu'_{k,m}$ to denote the arm with the largest expected mixed reward for agent $m$, assumed unique. Also, for all $m$ and $k$, we use $\Delta'_{k,m}=\mu'_{k^*_m,m}-\mu'_{k,m}$ to denote the gap between the expected mixed reward of arm $k$ and the arm with the highest mean mixed reward, for agent $m$. 
In addition, for all $m$ and $k\neq k^*_m$, we define $\Deltat'_{k, m}=\Delta'_{k,m}$. In the case of the best arm $k^*_m$ of each agent $m$, we define $\Deltat'_{k^*_m,m}=\min_{k\neq k^*_m}\Delta'_{k,m}$, equal to the gap for the second best arm. We introduce the notation $\Deltat'$ that is different from $\Delta'$ only in the case of the best arm, for the convenience of presenting our results and proofs.
We also use the notations $\Delta'_{\min} = \min_{k,m}\Deltat'_{k,m}$ and $\Delta'_{\max} = \max_{k, m}\Deltat'_{k,m}$ to denote the smallest and largest gaps for agents, respectively.

At each time $t$, the agents are allowed to communicate information to a server, in the form of empirical mean rewards of local observations~\citep[similar to][]{ reda2022collaborative,shi2021federated}. The information is then broadcast to all other agents. We aim at designing efficient algorithms with as few as possible communication rounds. A collaborative learning bandit algorithm $\pi=\{\pi_m\}_{m\in[M]}$ is a specification of the arm $\pi_{m}(t)\in[K]$ selected by agent $m$ at time $t$. The performance of $\pi$ is measured in terms of its collaborative regret defined as follows:
\begin{equation}
 \Rc(T) = \E\left[\sum_{m=1}^M\sum_{t=1}^T(\mu'_{k^*_m,m} - \mu'_{\pi_m(t),m})\right].
\end{equation}
\cite{shi2021federated} considered a special case of the collaborative bandit model with $w_{m,m}=\alpha+\frac{1-\alpha}{M}$ and $w_{n,m}=\frac{1-\alpha}{M}$ for $n\neq m$, for some $\alpha\in[0,1]$. We note that when $w_{m,m}=1$ for all $m\in[M]$ the problem reduces to the degenerated case of $M$ independent single-agent bandit problems. Moreover, when $w_{n,m}=\frac{1}{M}$ for all $n,m\in[M]$, the problem reduces to another special case where all agents are interested in a global average of the rewards across all agents considered in \cite{shi2021global}.
%~\citep[the setting considered in][]{shi2021global}.  

We use the following notations for the sample means. 
The sample mean of local observations of arm $k$ by agent $m$ is denoted by
\begin{equation}
 \hat{\mu}_{k,m}(t)=\frac{1}{\tau_{k,m}(t)}\sum_{s=1}^tX_{k,m}(s)\one\{\pi_m(s)=k\},
\end{equation}
where $\tau_{k, m}(t)$ denotes the number of times arm $k$ was selected by agent $m$ by the end of round $t$, and $\one$ is the indicator function, i.e.\ for an event $\Fc$, $\one\{\Fc\}=1$ if $\Fc$ holds true, and $\one\{\Fc\}=0$, otherwise. 
% Also, let us use the notation $\not~\Fc$ for the complement of $\Fc$.
The sample mean of the mixed reward is denoted by
\begin{align}
    \hat{\mu}'_{k, m}(t)=\sum_{n=1}^Kw_{n,m}\hat{\mu}_{k,n}(t).
\end{align}

\subsection{Regret Lower Bound}

\cite{reda2022collaborative} established a regret lower bound for the collaborative bandit model that also confirmed the conjecture in~\cite{shi2021federated} for their special case of weights. This result is formally stated in the following lemma. The lower bound holds for normal distributions over the rewards. Furthermore, the agents are allowed to communicate after each round. 
\begin{lemma}\label{lem:lowerbound}
Any uniformly efficient collaborative bandit algorithm
in which all agents communicate after each round
satisfies $\liminf_{T\rightarrow \infty}\frac{\Rc(T)}{\log(T)}\geq c^*$, where
\begin{align}
&c^*= \min_{q\in(\Rr^+)^{K\times M}} \sum_{m\in[M]}\sum_{k\in[K],k\neq k^*_m}q_{k,m}\Deltat'_{k,m}, \\
&\text{subject to:}~ \forall m\in[M], \forall k\in[K],\ \sum_{\{n\in[M]: k^*_n\neq k\}}\frac{w^2_{n,m}}{q_{k,n}}\le\frac{(\Deltat'_{k,m})^2}{2}.
\end{align}
\end{lemma}
The proof is based on the change of measure technique used in proving regret lower bounds in the single-agent bandit problem~\citep[see,][]{LaiRobbins85bandits, salomon2011regret}.
A uniformly efficient algorithm satisfies $\Rc(T)={o}(T^\gamma)$ for any $\gamma\in(0,1)$ and for all possible distribution instances $\nu=\{\nu_{k,m}\}_{k\in[K],m\in[M]}$.
The lower bound complexity term $c^*$ can be intuitively understood as the value of regret divided by $\log(T)$ for the optimum allocation of the arm plays, guaranteeing sufficiently small confidence intervals for all mean mixed rewards. The intuition is based on the observation that the term $\sum_{\{n\in[M]: k^*_n\neq k\}}\frac{w^2_{n,m}}{q_{k,n}}$ in the constraints in Lemma~\ref{lem:lowerbound} appears in the confidence interval width of the mean mixed rewards (see Equation~\ref{eq:omega}). 

A relaxed complexity term was given in~\cite{reda2022collaborative}, which does not require the knowledge of the best arm and is easier to work with in algorithm design. Let us define an oracle for the following optimization problem related to the lower bound, which will also be used in the algorithm.
\begin{definition}\label{def:P}
For any $\Delta\in(\Rr^+)^{K\times M}$, the oracle $\Pc(\Delta)$ is the solution to the following constrained optimization problem. 
\begin{align}
&\argmin_{q\in(\Rr^+)^{K\times M}} \sum_{k\in[K],m\in[M]}q_{k,m}\Delta_{k,m}, \\
&\text{subject to:}~ \forall m\in[M],\forall k\in[K],\ \sum_{n\in[M]}\frac{w^2_{n,m}}{q_{k,n}}\le\frac{\Delta^2_{k,m}}{2}.
\end{align}
\end{definition}
\cite{reda2022collaborative} proved that the complexity term $c^*$ in the regret lower bound in Lemma~\ref{lem:lowerbound} and a complexity term $\tilde{c}^*$ obtained from the relaxed oracle $\Pc$ are the same up to small constant factors.
\begin{lemma}\label{lem:relaxed}
Let $(q^*_{k,m})_{k,m}$ and $\tilde{c}^*$ be as $(q^*_{k,m})_{k,m} = \Pc(\Deltat')$ and $\tilde{c}^* = \sum_{k\in[K], m\in[M]}q^*_{k,m}\Deltat'_{k,m}$. Then it holds that
\begin{equation}
     c^*\le\tilde{c}^*\le 4c^*.
\end{equation}
\end{lemma}
We thus have $\frac{\tilde{c}^*}{4}\le c^*$, which implies, under the setting of Lemma~\ref{lem:lowerbound},
\begin{align}
    \liminf_{T\rightarrow \infty}\frac{\Rc(T)}{\log(T)}\geq \frac{\tilde{c}^*}{4}.
\end{align}

\subsection{Confidence Intervals}\label{sec:CI}

Confidence intervals serve as essential building blocks in the design of efficient bandit algorithms. For any $k\in[K], m\in[M]$, $\delta \in (0, 1)$, at time $t$, we define:
    \begin{equation}\label{eq:omega}
        \Omega_{k, m}^\delta(t) \coloneqq \sqrt{\beta_{\delta}(\tau_{k, \cdot}(t)) \sum_{n=1}^M \frac{w_{n, m}^2}{\tau_{k, n}(t)}}
    \end{equation}
    where $\beta_\delta:(\Rr^+)^M\rightarrow \Rr^+$
    %$N \mapsto \beta_{\delta}(N)$ 
    is a threshold function defined as:
    \begin{align}
        \beta_{\delta}(N) \coloneqq 2\left(g_M\left(\frac{\delta}{KM}\right) + 2 \sum_{m = 1}^M \log(4 + \log(N_m))\right)
    \end{align}
    where $g_M$ is some non-explicit function, defined in \cite{kaufmann2018mixture} that satisfies:
    \begin{align}
        g_M(\delta) \simeq \log(\frac{1}{\delta}) + M\log\log(\frac{1}{\delta})
    \end{align}

We then have the following, uniform in time, confidence intervals for mixed mean rewards.  
\begin{lemma}\label{lem:confidencebound}\citep{kaufmann2018mixture, reda2022collaborative}
For any $\delta \in (0, 1)$, the event $\Fc$ defined as follows holds with probability larger than $1-\delta$.
\begin{align}
    &\Fc \coloneqq \Bigg\{ \forall t \in \Nn, \forall k\in[K], m\in[M]; 
    ~|\hat{\mu}'_{k, m}(t) - \mu'_{k, m}| \leq \Omega_{k, m}^\delta(t) \Bigg\}
\end{align}
\end{lemma}

Lemma~\ref{lem:confidencebound} is proven in~\citep[][Proposition $24$]{kaufmann2018mixture} by constructing mixture martingales resulting in tight confidence intervals for certain functions of the means of the arms. We note that, in our setting, the standard Chernoff-Hoeffding type inequalities cannot be directly applied, as we need to create confidence intervals for $\mu'_{k,m}$ utilizing observations of $X_{k,n}$, $n\in[M]$, in contrast to directly observing samples form $X'_{k,m}$.

\section{Collaborative Learning in Bandits: A Regret Minimization Algorithm}\label{sec:alg}

In this section, we present \alg, a collaborative bandit algorithm for regret minimization. The existing work has primarily focused on phased elimination algorithms, which seems a natural choice in settings like collaborative bandits. These algorithm designs are motivated by the lower bound. During each phase, the lower bound oracle is employed to obtain a good allocation of the arm plays based on the current estimate of the problem parameters (here the gaps $\Deltat'_{k,m}$).

The allocation obtained by the oracle is then used for a guided exploration of arms in the next phase. Suboptimal arms are identified and removed from the set of candidate optimal arms at the end of each phase. In the best arm identification setting, \cite{reda2022collaborative} introduced W-CPE-BAI, a phased elimination algorithm that achieves near-optimal sample complexity. The performance of a similar algorithm in the regret minimization setting, however, does not match the regret lower bound, as discussed in~\cite{reda2021dealing}.

A similar shortcoming is also observed for PF-UCB introduced in~\cite{shi2021federated} for regret minimization in the special weights setting. The reason for this drastic difference between best arm identification and regret minimization settings is the subtle difference in the constrained optimization problems associated with the lower bound complexities in each setting. In particular, we note that the unknown gaps $\Deltat'_{k,m}$ appear both in the objective as well as in the constraints in the Oracle $\Pc$ given in Definition~\ref{def:P}. 

This is in contrast to the constrained optimization problem associated with the lower bound on the sample complexity where the unknown gap parameters appear only in the constraints~\citep[see,][Theorem~$1$]{reda2022collaborative}. This subtle difference breaks the analysis of standard phased elimination algorithms for regret minimization in the collaborative bandit model and leaves a significant gap in the performance~\citep{reda2022collaborative}. To address this problem, we introduce \alg (see Algorithm~\ref{alg:1}).

\subsection{Description of \texorpdfstring{\alg~Algorithm}{algorithm}}\label{sec:alg-desc}

\alg~comprises two exploration phases followed by a third phase in which all agents either exploit ---by playing the arm with the highest sample mean of mixed reward--- or switch to any policy $\pi'$ with an $\Oc(\log(T))$ regret.
%(not necessarily with optimal constant in front of the log term). 

An example of policy $\pi'$ with logarithmic regret is a variation of W-CPE-BAI adopted to the regret minimization setting. We formally give this variation, 
%referred to as 
W-CPE-Reg, and the proof that it has an $\Oc(\frac{c^*}{\Delta'_{\min}}\log(T))$ regret using an $\Oc(\log(\frac{1}{\Delta'_{\min}}))$ communication rounds, in Section~\ref{sec:WCPEReg}. 

\paragraph{Initial Exploration:} In the initial sub-logarithmic exploration phase of \alg, each agent $m\in [M]$ plays each arm $k\in [K]$ for $\tau_1 = \lceil\sqrt{\log(T)}\rceil$ times. The length of the initial exploration phase is denoted by 
\begin{align}
    T_{\ie}=MK\tau_1.
\end{align}
At the end of the initial exploration phase,  each agent $m$ computes their local sample mean reward $\hat{\mu}_{k,m}(T_{\ie})$ for each arm $k$. The sample means of local rewards are shared through the server with other agents (the first round of communication), which are used to compute the sample means of mixed rewards $\hat{\mu}'_{k,m}(T_{{\ie}})$, and estimates of gaps in mixed rewards $\hat{\Delta}'_{k,m}(T_{\ie})$. 

\paragraph{Guided Exploration:}
\alg~utilizes the gap estimates in mixed rewards and Oracle $\Pc$ to obtain an allocation of the arm plays $(\tau^{\ge}_{k,m})_{k,m}$ for the guided exploration phase, and ensures each agent $m$ plays each arm $k$ until reaching $\tau^{\ge}_{k,m}$.
To avoid extreme cases of allocation, which may lead to large regret, the gap estimates are first projected onto the $(\frac{1}{\log\log(T)},\log\log(T))$ interval. Specifically, we introduce, $\forall k,m$,
\begin{align}
    \begin{split}
        &\Deltah_{k,m}^{\ge} = \min\Bigg\{\max\left\{\Deltah'_{k,m}(T_{\ie}), \frac{1}{\log\log(T)}\right\},
        ~\log\log(T)\Bigg\}.
    \end{split}
\end{align}
Then, the allocation design $q=\Pc(\Deltah^{\ge})$ is obtained using $\Deltah^{\ge}$. The allocation of the arm plays in the guided exploration phase is then set to 
$\lceil18q_{k,m}^{\ge}B(T)\rceil$,
where 
$
B(T) = \beta_{\delta'}(\tau_{k,\cdot})
$, with $\delta'=\frac{1}{T}$ and $\tau_{k,n}=T$, $\forall k,n$, is an upper bound on $\beta_{\delta'}(\tau_{k,\cdot})
$. As given in Section~\ref{sec:CI}, 
\begin{align}
\begin{split}
    B(T) \simeq &~2\log(KMT) + 2M\log\log(KMT) + 4M\log(4+\log(T)).
\end{split}
\end{align}
Thus, we have $B(T)=\Oc(\log(KMT))$ hiding a small absolute constant $2$ for large $T$. In the guided exploration phase, each agent $m$ plays each arm $k$ to reach a total number of $\tau^{\ge}_{k,m}=\max\{\tau_1, \lceil18q_{k,m}^{\ge}B(T)\rceil\}$. Let $T_{\ge}$ denote the end of guided exploration phase. At this time, the sample means of local rewards are shared through the server with other agents (the second
round of communication). Those are used to compute $\hat{\mu}'_{k,m}(T_{\ge})$ and $\Deltah_{k,m}'(T_{\ge})$ similar to the end of initial exploration phase. 

\paragraph{Exploitation/ Switch to $\pi'$:} 
At the end of the guided exploration phase, \alg~either exploits ---by playing the arm with the highest sample mean of mixed reward--- or switches to $\pi'$, based on the following condition, with $\delta'=\frac{1}{T}$,
\begin{align}\label{eq:cond}
\begin{split}
    &\Cc=\Bigg\{\forall k \in [K],m \in [M],
    ~\Omega^{\delta'}_{k,m}(T)< \frac{\Deltah'_{k,m}(T_{\ge})}{2}\Bigg\}.
\end{split}
\end{align}
If $\Cc$ does not hold, \alg~switches to $\pi'$. Otherwise, at the end of the guided exploration phase, each agent $m$ identifies the arm with the largest empirical mean mixed reward 
\begin{equation}
\hat{k}_m=\argmax_{k\in[K]}\hat{\mu}'_{k,m}(T_{\ge}),
\end{equation}
and plays this arm for the remaining time. 
A detailed pseudo-code for \alg~is given in Algorithm~\ref{alg:1}. The regret performance and the number of communication rounds of \alg~are presented in the following theorem and lemma.   

\begin{algorithm}[ht]
\caption{\alg, regret minimization in collaborative bandit model.}
\label{alg:1}
\begin{algorithmic}
\STATE Input: $M$ agents, $K$ arms, weight matrix $W$, $\pi'$

\STATE \textbf{Initial Exploration:} 
\STATE Each agent $m$, plays each arm $k$ for $\tau_1=\lceil\sqrt{\log(T)}\rceil$ times.
\STATE  Agents communicate the sample mean of their local observations.
\STATE  Agents obtain $(\Deltah'_{k,m}(T_{\ie}))_{k,m}$.

\STATE \textbf{Guided Exploration:}
\STATE  $\Deltah^{\ge}_{k, m} \gets \min\left\{\max\{\Deltah'_{k, m}(T_{\ie}), \frac{1}{\log\log(T)}\},\log\log(T)\right\}$, $\forall k \in [K], m \in [M]$
\STATE  Let $(q^{\ge}_{k,m})_{k, m}=\Pc(\Deltah^{\ge})$
%\WHILE{$t \leq \max_{k, m}\{ \lceil18q_{k, m}^{\ge} B(T)\rceil\}$ }
\STATE  Each agent $m \in [M]$ plays each action $k \in [K]$ until it reaches to a total number of \begin{equation*}
    \tau^{\ge}_{k,m}=\max\{\tau_1, \lceil18q_{k,m}^{\ge}B(T)\rceil\}
\end{equation*}
\STATE  Agents communicate the sample mean of their local observations
\STATE  Agents update $(\Deltah'_{k,m}(T_{\ge}))_{k,m}$

\STATE \textbf{Exploitation/Switch to {$\pi'$}:}
\IF{$\forall k \in [K],m \in [M],~\Omega^{\delta'}_{k,m}(T)< \frac{\Deltah'_{k,m}(T_{\ge})}{2}$}
\STATE  Each agent $m$ plays $\argmax_{k\in[K]}\hat{\mu}'_{k,m}(t)$ until $T$
\ELSE
\STATE  Reset all data and play based on $\pi'$ until $T$
\ENDIF
\end{algorithmic}
\end{algorithm}

\begin{theorem}\label{the:main}
Consider the collaborative bandit model described in Section~\ref{sec:PF}. Consider \alg~given in Algorithm~\ref{alg:1}. Let $\pi'$ be W-CPE-Reg. The regret performance of \alg~satisfies, for some $T_0\in\Nn$, and for all $T\geq T_0$,
\begin{equation}
\Rc(T) =\Oc\left(c^*\log(T)
+\frac{(\Delta'_{\max})^2}{\Delta'_{\min}}(\log\log(T))^4\right)
\end{equation}
\end{theorem}

The $\Oc$ notation hides only absolute constants which are specified in the analysis section. The value of the constant $T_0$ is also given in the analysis section. 
The $\Oc(c^*\log(T))$ regret bound given in Theorem~\ref{the:main} proves that the performance of \alg~matches the lower bound up to an absolute constant factor. As mentioned, \alg~also enjoys very few expected number of communication rounds, as given in the following lemma. 

\begin{lemma}\label{lemma:comm}
Under the setting of Theorem~\ref{the:main},
the expected number of communication rounds in \alg~is bounded by 
\begin{align}
    2 +{\lceil \log_2(\frac{8}{\Delta'_{\min}}) \rceil}/{\log(T)}.
\end{align}
In addition, with probability at least $1-\frac{1}{\log(T)}$ the number of communication rounds in \alg~is exactly~$2$. 
\end{lemma}

According to Lemma~\ref{lemma:comm}, as $T$ grows large, \alg~requires only $2$ communication rounds. In comparison, W-CPE-BAI required $\lceil \log_2(\frac{8}{\Delta'_{\min}}) \rceil$ communication rounds. The proof is provided in Appendix~\ref{app:comm}.

\subsection{An Example of Algorithms with Logarithmic Regret \texorpdfstring{($\pi'$)}{pi-prime}}\label{sec:WCPEReg}

For completeness and correctness of our results, we provide a concrete example of a collaborative bandit algorithm with logarithmic regret 
(not necessarily with optimal constant in front of log term), 
that can be used as $\pi'$ in \alg. This algorithm, referred to as W-CPE-Reg, is similar to W-CPE-BAI~\citep{reda2022collaborative}, but adopted to the regret minimization setting. W-CPE-BAI was originally introduced %in~\cite{reda2022collaborative} 
for best arm identification, where the goal of each agent $m$ is to return their candidate best arm $\hat{k}_m$, with as few samples from arms as possible, satisfying, for a given $\delta\in(0,1)$,
\begin{equation}
    \Pr\left[ \forall m\in[M], \hat{k}_m=k^*_m\right]\geq 1-\delta.
\end{equation}
We run W-CPE-BAI with $\delta=\frac{1}{T}$ until it stops. Then, each agent $m$ plays $\hat{k}_m$ for the remaining time. Referring to this procedure as W-CPE-Reg, we have the following result on its regret performance and communication rounds. The proof is provided in Appendix~\ref{app:W-CPE-Reg}.
\begin{lemma}\label{lem:W-CPE-Reg}
Consider the collaborative bandit model described in Section~\ref{sec:PF}. The number of communication rounds is at most $\lceil \log_2(\frac{8}{\Delta'_{\min}}) \rceil$. Moreover, the regret performance of W-CPE-Reg described above satisfies
\begin{equation}
    \Rc(T) \le \frac{128c^*}{\Delta'_{\min}}\log_2(\frac{8}{\Delta'_{\min}})\log(T) + o(\log(T)).
\end{equation}
\end{lemma}

\section{Analysis}\label{sec:anal}

In this section, we overview the main steps and challenges in the proof of Theorem~\ref{the:main}, while deferring some details and proof of lemmas to the appendix. 
The regret of \alg~is partitioned into three disjoint parts: the regret $\Rc_{\ie}$ during the initial exploration phase $t\in \Tc_{\ie}$, the regret $\Rc_{\ge}$ during the guided exploration phase $t\in \Tc_{\ge}$, and the regret $\Rc_{\ex}$ during the third phase $t\in \Tc_{\ex}$ (exploitation or switch to $\pi'$), which we refer to as exploitation for brevity. We note that
\begin{align}
    \Rc(T)=\Rc_{\ie} + \Rc_{\ge} +\Rc_{\ex}.
\end{align}

Let us first introduce two high probability events, $\EE_T$ and $\EB_T$, on the sample means of mixed rewards with $\delta=\frac{1}{\log(T)}$ and $\delta'=\frac{1}{T}$. 
\begin{align}
    &\EE_T = \Bigg\{ \forall t \leq T, \forall k\in[K], m\in[M]; 
    \|\hat{\mu}'_{k, m}(t) - \mu'_{k, m}\| \leq \Omega_{k, m}^\delta(t) \Bigg\}, \\
    &\EB_T = \Bigg\{ \forall t \leq T, \forall  k\in[K], m\in[M]; 
    \|\hat{\mu}'_{k, m}(t) - \mu'_{k, m}\| \leq \Omega_{k, m}^{\delta'}(t) \Bigg\}.
\end{align}

Based on Lemma~\ref{lem:confidencebound}, we have $\Pr[\EE_T]\geq 1-\frac{1}{\log(T)}$ and $\Pr[\EB_T]\geq 1-\frac{1}{T}$. These two events are used in the analysis of $\Rc_{\ge}$ and $\Rc_{ex}$. Also, for the sake of brevity, we use the notations $\delta=\frac{1}{\log(T)}$ and $\delta'=\frac{1}{T}$ throughout the proof. We now bound $\Rc_{\ie}$, $ \Rc_{\ge}$ and $\Rc_{\ex}$ to prove the theorem.
\paragraph{Regret in Initial Exploration:} In the initial exploration phase each agent $m\in[M]$ plays each arm $k\in[K]$ for $\tau_1=\lceil\sqrt{\log(T)}\rceil$ times. Thus, $\Rc_{ie}$ is simply bounded as follows
\begin{align}
\Rc_{ie} \leq KM\lceil\sqrt{\log(T)}\rceil \Delta'_{\max}. 
\end{align}

\paragraph{Regret in Guided Exploration:} 
By law of total expectation and using the notation ``$\text{not}~\Fc$'' for complement of an event $\Fc$, we have, $\Rc_{\ge}$ is equal to
\begin{align}
\begin{split}
    \E\left[ \sum_{t \in \Tc_{\text{ge}}}\sum_{m=1}^M \Delta'_{\pi_{m}(t), m} \right] &= \underbrace{\E \left[ \one\{\EE_T\}\sum_{t \in \Tc_{\text{ge}}} \sum_{m=1}^M \Delta'_{\pi_{m}(t), m} \right]}_{\textcolor{blue}{\text{Term}~1}}\\
     &\qquad \qquad \qquad \qquad + \\
     &\quad \underbrace{\E \left[ \one\{\text{not }\EE_T\}\sum_{t \in \Tc_{\text{ge}}} \sum_{m=1}^M \Delta'_{\pi_{m}(t), m} \right]}_{\textcolor{blue}{\text{Term}~2}}.
\end{split}
\end{align}

We proceed by bounding $\Rc_{\ge}$ when $\EE_T$ holds true and when it does not. 

\paragraph{$\EE_T$ holds true.}
We first show that $\Deltah^{\ge}$ used in designing the allocation in the guided exploration phase are within constant factors of the gaps $\Deltat'$.
Recall $\Omega_{k,m}^{\delta}(t)$ given in~\eqref{eq:omega}. At the end of the initial exploration phase, we have 
\begin{align}
        \Omega^{\delta}_{k,m}(T_{\ie})
        &= \sqrt{2\left(g_M(\frac{\delta}{KM})+2M\log(4+\log(\tau_1))\right)\sum_{n=1}^M \frac{w^2_{n,m}}{\tau_1}}\\
    &=\Oc(\sqrt{ \frac{\log (KM)+\log\log(T) +\log\log(KM\log(T))}{\sqrt{\log(T)}} }),
\end{align}
where $\Oc$ notation hides only small absolute constants. Note that 
\begin{align}
    \sum_{n=1}^M w^2_{n,m}\le \sum_{n=1}^M w_{n,m}=1.
\end{align}
Therefore, $\Omega^{\delta}_{k,m}(T_{\ie})$ is decreasing in $T$ as given above. Thus, for a sufficiently large $T_1$, for all $T\geq T_1$ and $\forall k, m$, we have 
\begin{align}
    \Omega^{\delta}_{k,m}(T_{\ie})\le \frac{\Delta'_{k,m}}{4}.
\end{align}
From the confidence bounds on the sample means of mixed rewards, conditioned on $T\geq T_1$ and $\EE_T$, we have that
\begin{align}
    \|\Deltah'_{k, m}(T_{\ie}) - \Deltat'_{k, m}\| \leq \frac{\Deltat'_{k, m}}{2}.
\end{align}
Thus $\Deltah'$ are the same as $\Deltat'$ up to absolute constants. \emph{I.e.}\ $\frac{1}{2}\Delta'_{k, m} \leq \Deltah'_{k, m} \leq \frac{3}{2}\Delta'_{k, m}$, for all $k$, $m$.

In addition, let $T_2$ and $T_3$ be the smallest numbers for which $\frac{1}{\log\log(T_2)} \leq \frac{1}{2}\Delta'_{k,m}$, for all $k,m$, and $\frac{3}{2}\Delta'_{k,m} \leq \log\log(T_3)$, for all $k,m$, respectively. Let $T_0=\max\{T_1, T_2, T_3\}$.

Recall the projections $\Deltah_{k,m}^{\ge}$ of $\Deltah'_{k,m}(T_{\ie})$ onto $(\frac{1}{\log\log(T)},\log\log(T))$ interval. Under event $\EE_T$, when $T\geq T_0$, we have $\Deltah_{k,m}(T_{\ie})=\Deltah'^{\ge}_{k,m}$. Thus,
\begin{eqnarray}\label{eq:const}
     \frac{1}{2}\Delta'_{k, m} \leq \Deltah_{k,m}^{\ge}\leq \frac{3}{2}\Delta'_{k, m}
\end{eqnarray}
The following Lemma shows that the complexity term in the oracle $\Pc$ remains within constant factors, when the input is scaled by up to constant factors. The proof is provided in Appendix \ref{app:cfac}.
\begin{lemma}\label{lem:cfac}
    Recall Oracle $\Pc$ from Definition~\ref{def:P}. Consider two sets of gaps $\Delta, \Deltah\in(\Rr^+)^{K\times M}$. Let $q^* =\Pc(\Delta)$, $\hat{q}^* =\Pc(\Deltah)$,
    and $c = \sum_{k\in[K],m\in[M]}q_{k,m}\Delta_{k,m}$,$\hat{c} = \sum_{k\in[K],m\in[M]}\hat{q}_{k,m}\Delta_{k,m}$.
    Assume $\forall k\in[K], m\in[M]$, $a\Delta_{k,m}\le \Deltah_{k,m}\leq b\Delta_{k,m}$, for some $0<a<b$. 
    Then, $\hat{c} \le \frac{b}{a^3}c$. 
\end{lemma}

Using Lemma~\ref{lem:cfac}, the regret in the guided exploration phase, under event $\EE_T$ and assuming that $T\geq T_0$, can be bounded as
\begin{align}
    \text{Term}~1
    &\le \sum_{m\in[M], k\in[K]}\lceil18q^{\ge}_{k,m}B(T)\rceil\Delta'_{k,m}\\
    &\le
    18\sum_{m\in[M], k\in[K]}(q^{\ge}_{k,m}\Deltat'_{k,m})B(T) + \sum_{m\in[M], k\in[K]} \Delta'_{k, m} \\
    &\le 216\tilde{c}^*B(T) + KM\Delta'_{\max},
\end{align}
where the first inequality comes from $\Pr[\EE_T]\le 1$ and the third inequality holds by Lemma~\ref{lem:cfac} and~\eqref{eq:const}. We  note that $\frac{B(T)}{\log(T)}\simeq 2$ for large $T$ by definition. Term $1$ is the dominant term in the regret bound.

\paragraph{{not}~$\EE_T$ holds true.} To upper bound Term $2$, let $a = \frac{1}{\Delta'_{\max}\log\log(T)}$ and $b = \frac{\log\log(T)}{\Delta'_{\min}}$. For $T \geq T_0$, \eqref{eq:const} implies
\begin{equation}
    a\Delta'_{k, m} \leq \Deltah'_{k, m} \leq b\Delta'_{k, m} \quad \forall k \in [K], m\in [M].
\end{equation}
Using Lemma \ref{lem:cfac}.
\begin{align}
    \text{Term}~2
    &= \Pr(\text{not }\EE_T)\sum_{m\in[M], k\in[K]}\lceil18q^{\ge}_{k,m}B(T)\rceil\Delta'_{k,m}\\
    &\le \Pr(\text{not }\EE_T) \Bigg(
    18\sum_{m\in[M], k\in[K]}(q^{\ge}_{k,m}\Deltat'_{k,m})B(T) + \sum_{m\in[M], k\in[K]} \Delta'_{k, m}\Bigg) \\
    &\leq \frac{1}{\log(T)}\Bigg(18\left(\log\log(T)\right)^4 \frac{(\Delta'_{\max})^3}{\Delta'_{\min}} \tilde{c}^*B(T) + KM\Delta'_{\max}\Bigg),
\end{align}
where the last inequality comes from Lemma~\ref{lem:cfac} and $\Pr[\text{not}~\EE_T]\le \frac{1}{\log(T)}$. Since $\frac{B(T)}{\log(T)}\simeq 2$ for large $T$, we have
\begin{align}
    \text{Term}~2 =\Oc\left(\left(\log\log(T)\right)^4 \frac{(\Delta'_{\max})^3}{\Delta'_{\min}} \tilde{c}^*\right),
\end{align}
up to an absolute constant factor of approximately $36$. 

\paragraph{Regret in Exploitation/Switch to $\pi'$:}
We first consider the exploitation case where the condition $\Cc$ in~\eqref{eq:cond} is satisfied, and the algorithm proceeds with choosing $\hat{k}_m=\argmax_{k\in[K]}\hat{\mu}'_{k,m}(T_{\ge})$. When $\EB_T$ holds true, the best arm will be chosen.  Because, $\forall m,k\neq k^*_m$
\begin{align}
{\mu}'_{\hat{k}_m,m} &\geq~~
\hat{\mu}'_{\hat{k}_m,m}(T_{\ge})-\Omega^{\delta'}_{\hat{k}_m,m} \\
&>~~ \hat{\mu}'_{k_m,m}(T_{\ge})+\Omega^{\delta'}_{{k},m}\\
& \geq~~ {\mu}'_{k,m}. 
\end{align}

The second inequality holds by $\Cc$. 
Thus, there is no regret in the exploitation phase. When $\EB_T$ does not hold true, the regret can be simply bounded by $KM\Delta'_{\max}$, since $\Pr(\text{not}~ \EB_T)\le \frac{1}{T}$. 
Therefore, the expected regret under the exploitation case 
$\Rc_{\ex} \le KM\Delta'_{\max}$.

Now, for the case where $\Cc$ fails and the algorithm switches to $\pi'$, we have the following lemma.

\begin{lemma}\label{lem:events}
Recall $\EE_T$ and $\EB_T$ and condition $\Cc$. We have 
 \begin{equation}\label{eq:C}
     (\EE_T ~~\text{and}~~ \EB_T)  \implies \Cc.
 \end{equation}
\end{lemma}

Moreover, we note that ``not $\EE_T$'' implies ``not $\EB_T$'', since $\Omega_{k,m}^{\delta}\le  \Omega_{k,m}^{\delta'}$.
Thus, the probability of switching is bounded by the probability of ``not $\EE_T$'', which is at most $\frac{1}{\log(T)}$. 
Hence, the regret in this case is bounded by
\begin{equation}
\Rc_{\ex}\le \frac{\Rc^{\pi'}(T)}{\log(T)}\le \frac{128c^*}{\Delta'_{\min}}\log_2(\frac{8}{\Delta'_{\min}})+o(1). 
\end{equation}
 
Adding up the bounds on $\Rc_{\ie}$, $\Rc_{\ge}$ and $\Rc_{\ex}$ completes the proof. 

\section{Conclusion}\label{sec:conc}

In this work, we considered the collaborative bandit model given in~\cite{reda2022collaborative}. While near-optimal sample complexities were known under this model, the problem of optimal regret bounds was left as an open question. We addressed this problem and proposed a novel algorithm, \alg, achieving order optimal regret bounds by running two rounds of exploration. We also showed that \alg~needs only a small constant expected number of communication rounds. 

\newpage

\bibliography{references}

\newpage
\appendix

\section{Other related work}\label{appx:rw}

There is a vast literature on single-agent stochastic bandits under various statistical models and assumptions~\citep[see,][as some representative works]{LaiRobbins85bandits, auer2002finite, vakili2013deterministic, abbasi2011improved, srinivas2010gaussian, lattimore2020bandit}. In the classic setting, the performance of algorithms was primarily measured in terms of regret~\citep{LaiRobbins85bandits, auer2002finite}, while a best arm identification setting was also later considered in several works~\citep{EvenDaral06, gabillon2012best, Jamiesonal14LILUCB}. The collaborative bandit model formalized in \cite{shi2021federated, reda2022collaborative} is a generalization of the classic single-agent bandit to a multi-agent setting.

Collaborative bandit under a pure exploration setting, where all agents face the same bandit problem, has been considered in~\cite{hillel2013distributed, tao2019collaborative, karpov2020collaborative, chen2021cooperative, zhu2021decentralized}. Some of these works consider a communication matrix where agents are allowed to communicate only with their neighbors~\citep[see,][]{zhu2021decentralized}. 
The goal of collaborative pure exploration is to reduce sample complexity at the cost of communication rounds. \cite{du2021collaborative} considered a collaborative bandit setting where a kernel is used to model the reward means. They provided a lower bound on the sample complexity of the best arm identification and also an algorithm with order optimal performance.
\cite{salgia2023collaborative} considered regret minimization under a collaborative kernel-based setting.
The kernel model, however, is inherently different from the multi-armed bandit considered in the collaborative learning model of~\cite{reda2022collaborative}.

The phased elimination algorithms are inspired by a simple phased elimination algorithm introduced in~\cite{auer2010ucb} for subtle improvements in the regret bounds of the classic single-agent bandit. These algorithms maintain a set of potential optimal arms which are pruned at the end of each phase based on the observations gathered so far. This technique has been used in several bandit algorithms in various settings for both regret minimization and best arm identification~\citep{hillel2013distributed, chen2017nearly, fiez2019sequential, BKMP20, shi2021federated, reda2022collaborative, du2021collaborative}. Recently, \cite{jin2021double} studied the standard $K$-armed bandits using a double-exploration algorithm. However, the algorithm in \cite{jin2021double} follows an ``explore-commit-explore-commit" pattern, while none of the exploration phases are guided in the sense of our guided exploration phase. Our algorithm follows an ``explore-guided explore-exploit/switch" phases, where in addition to the other difference mentioned before, the last phase includes switching to a policy which could be a UCB-based policy instead of always committing to one arm, which is suboptimal. Moreover, our multi-agent setting includes a constrained optimisation problem governing the lower bound which makes the problem significantly different than the setting in \cite{jin2021double}. In addition, our model due to its collaborative nature requires a more complex confidence bound.

\section{Proof of Lemma~\ref{lemma:comm}}\label{app:comm}

In \alg, agents communicate with the server at the end of both exploration phases, i.e.\ initial exploration and guided exploration. If the condition \eqref{eq:cond} holds, the algorithm enters the exploitation phase where no communication happens. Otherwise, the algorithm switches to $\pi'$, where the agents communicate according to $\pi'$. Therefore, the expected number of communication rounds is bounded by
\begin{equation}
     2 + \lceil\log_2(\frac{8}{\Delta'_{\min}}) \rceil\Pr(\text{not }\Cc),
\end{equation}
where $ \lceil \log_2(\frac{8}{\Delta'_{\min}}) \rceil$ is the number of communication rounds in $\pi'$ introduced is Section \ref{sec:WCPEReg}.

\section{Proof of Lemma~\ref{lem:W-CPE-Reg}}\label{app:W-CPE-Reg}

\cite{reda2022collaborative} considered the best arm identification setting. They introduced the following relaxed complexity term for the sample complexity lower bound. 
\begin{align*}
&s^*= \min_{q\in(\Rr^+)^{K\times M}} \sum_{m\in[M]}\sum_{k\in[K]}q_{k,m}, \\
&\text{subject to:}~ \forall m\in[M], \forall k\in[K],~\sum_{\{n\in[M]: k^*_n\neq k\}}\frac{w^2_{n,m}}{q_{k,n}}\le\frac{(\Delta'_{k,m})^2}{2}.
\end{align*}

They proved that the sample complexity of W-CPE-BAI is at most
\begin{equation}
    32s^*\log_2(\frac{1}{\Delta_{\min}})\log(\frac{1}{\delta})+o(\log(\frac{1}{\delta})).
\end{equation}

Choosing $\delta=\frac{1}{T}$, we have that the regret of W-CPE-BAI satisfies
\begin{align}
\begin{split}
    \Rc(T) \le& 32s^*\log_2(\frac{1}{\Delta_{\min}})\log(T) + o(\log(T)) + \frac{TM\Delta_{\max}}{T}.
\end{split}
\end{align}

In addition, we prove that $s^*\le \frac{4c^*}{\Delta'_{\min}}$ that completes the proof. Let $\hat{q}^*$ be the $\argmin$ in the above optimization problem. Also, let ${q}^*$ be the $\argmin$ in the optimisation problem of Oracle $\Pc$ given in Definition~\ref{def:P} and $\tilde{c}^*$ be as in Lemma~\ref{lem:relaxed}. By definition 
\begin{eqnarray}\nn
s^* &=& \sum_{m\in[M],k\in[K]}\hat{q}^*_{k,m} \\\nn
&\le&\sum_{m\in[M],k\in[K]}{q}^*_{k,m} \\\nn
&\le& \frac{1}{\Delta'_{\min}} \sum_{m\in[M],k\in[K]}{q}^*_{k,m}\Delta'_{k,m}\\\nn
&=&\frac{\tilde{c}^*}{\Delta'_{\min}}\\\nn
&\le&\frac{4{c}^*}{\Delta'_{\min}}.
\end{eqnarray}
The last inequality holds by Lemma~\ref{lem:relaxed}. Also, note that the number of communication rounds follows the best arm identification setting, which is proved in \cite{reda2022collaborative} to be at most $\lceil \log_2(\frac{8}{\Delta'_{\min}}) \rceil$.

\section{Proof of Lemma~\ref{lem:cfac}}\label{app:cfac}

We note that, $\forall m\in[M]$,
$\sum_{n\in[M]}\frac{a^2w^2_{n,m}}{q_{k,n}}\leq \frac{a^2\Delta^2_{k,m}}{2}\leq \frac{\Deltah^2_{k,m}}{2}$.
This shows that $\{\frac{q_{k,m}}{a^2}\}_{k,m}$ satisfy the constraints in $\Pc(\Deltah)$. Thus, by definition 
\begin{eqnarray}\label{eq:ax1}
\sum_{k\in[K], m\in[M]}\hat{q}_{k,m}\Deltah_{k,m} \le \sum_{k\in[K], m\in[M]}\frac{{q}_{k,m}}{a^2}\Deltah_{k,m}.
\end{eqnarray}

Then, we have
\begin{eqnarray}\nn
\hat{c} &=& \sum_{k\in[K], m\in[M]}\hat{q}_{k,m}\Delta_{k,m}\\\nn 
&\le&\frac{1}{a} \sum_{k\in[K], m\in[M]}\hat{q}_{k,m}\Deltah_{k,m}\\\nn
&\le& \frac{1}{a^3} \sum_{k\in[K], m\in[M]}{q}_{k,m}\Deltah_{k,m}\\\nn
&\le& \frac{b}{a^3} \sum_{k\in[K], m\in[M]}{q}_{k,m}\Delta_{k,m}\\\nn
&=&\frac{b}{a^3}c
\end{eqnarray}
The first and third inequalities follows from $a\Delta_{k,m}\le \Deltah_{k,m}\le b\Delta_{k,m}$. The second inequality was shown in~\eqref{eq:ax1}. 

\section{Proof of Lemma~\ref{lem:events}}

Assume $\EE_T$ and $\EB_T$ hold true.  Recall the Oracle $\Pc$ from Definition~\eqref{def:P}. For $\tau^{\ge}$, the number of samples in the guided exploration phase, we have
\begin{eqnarray}\nn
\Omega_{k,m}^{\delta'}(T) &\le& \sqrt{\beta_{\delta'}(\tau_{k,\cdot}(T))\sum_{n=1}^M\frac{w^2_{n,m}}{\tau^{\ge}_{k,n}}}\\\nn
&\le& \sqrt{\beta_{\delta'}(\tau_{k,\cdot}(T))\sum_{n=1}^M\frac{w^2_{n,m}}{\lceil18q_{k, n}^{ge} B(T)\rceil}}\\\nn
&\le& \sqrt{\frac{(\Deltah^{\ge}_{k,m})^2}{36}}\\\label{eq:axx1}
&\le& \frac{\Delta'_{k,m}}{4}.
\end{eqnarray}
The third inequality holds by definition of $q^{\ge}$ and the constraints in Oracle $\Pc$. The last inequality holds by~\eqref{eq:const}.  
Under $\EB_T$, we also have, $\forall k,m$  
\begin{eqnarray}\nn
\|\hat{\mu}_{k,m}(T_{\ge})-\mu'_{k,m}\|&\le& \Omega_{k,m}^{\delta'}(T).
\end{eqnarray}

That implies, $\forall k,m$,
\begin{align}\label{eq:axx2}
\Deltah'_{k,m}(T_{\ge}) \geq \Delta'_{k,m}-2\Omega_{k,m}^{\delta'}(T) \geq \frac{\Delta'_{k,m}}{2}.
\end{align}

From~\eqref{eq:axx1} and~\eqref{eq:axx2}, we conclude, $\forall k,m$,
\begin{equation}
    \Omega_{k,m}^{\delta'}(T) \le \frac{\Deltah'_{k,m}}{2}.
\end{equation}

We thus proved that if both $\EE_T$ and $\EB_T$ hold true, then condition $\Cc$ holds true.

\end{document}